Enhancing Surgical Performance in Cardiothoracic Surgery with Innovations from Computer Vision

and Artificial Intelligence.

A Narrative Review.

Merryn D. Constable,[1] Hubert P. H. Shum,[2] & Stephen Clark[3,4]

[1]Department of Psychology, Northumbria University, Newcastle-upon-Tyne, UK.

[2]Department of Computer Science, Durham University, Durham, UK.

[3] Department of Applied Sciences, Northumbria University, Newcastle-upon-Tyne, UK.

[4] Consultant Cardiothoracic and Transplant Surgeon, Freeman Hospital, Newcastle upon Tyne, UK

Corresponding Author:

Merryn Constable

Northumberland Building

College Lane

Newcastle upon Tyne

UK, NE1 8SG

merryn.constable@northumbria.ac.uk

Word Count: 6275

**Abstract**

When technical requirements are high, and patient outcomes are critical, opportunities for monitoring and improving surgical skills via objective motion analysis feedback may be particularly beneficial. This narrative review synthesises work on technical and non-technical surgical skills, collaborative task performance, and pose estimation to illustrate new opportunities to advance cardiothoracic surgical performance with innovations from computer vision and artificial intelligence. These technological innovations are critically evaluated in terms of the benefits they could offer the cardiothoracic surgical community, and any barriers to the uptake of the technology are elaborated upon. Like some other specialities, cardiothoracic surgery has relatively few opportunities to benefit from tools with data capture technology embedded within them (as with robotic-assisted laparoscopic surgery, for example). In such cases, pose estimation techniques that allow for movement tracking across a conventional operating field without using specialist equipment or markers offer considerable potential. With video data from either simulated or real surgical procedures, these tools can (1) provide insight into the development of expertise and surgical performance over a surgeon's career, (2) provide feedback to trainee surgeons regarding areas for improvement, (3) provide the opportunity to investigate what aspects of skill may be linked to patient outcomes which can (4) inform the aspects of surgical skill which should be focused on within training or mentoring programmes. Classifier or assessment algorithms that use artificial intelligence to 'learn' what expertise is from expert surgical evaluators could further assist educators in determining if trainees meet competency thresholds. With collaborative efforts between surgical teams, medical institutions, computer scientists and researchers to ensure this technology is developed with usability and ethics in mind, the developed feedback tools could improve cardiothoracic surgical practice in a data-driven way.

Cardiothoracic surgical performance depends on integrating an extensive body of knowledge with, often complex and nuanced, technical and non-technical skills [1]. Given that surgery occurs within the context of individual patients and environmental factors, understanding surgical expertise and performance in a meaningful way that informs patient care and surgical training is a particularly challenging problem. Investment in tools to objectively track and analyse human movement is commonplace in elite sports [2], and similar tools could be used in surgical environments to enhance performance and patient outcomes [3]. Movement tracking in a surgical setting is also not unusual, with performance metrics available through surgical techniques and procedures that support data capture [4,5]. However, it is relatively rare to analyse movement and technical expertise in open surgical procedures due to difficulties in extracting the required metrics. Yet, a data-driven approach that provides analytics linking intra-operative clinical and technical processes to patient outcomes may provide a means of targeted improvement in surgical care [6,7]. Recent computer vision innovations may open doors to similar tools that benefit the pursuit of cardiothoracic surgical excellence, which relies less heavily, in relative terms, on robotic and thoracoscopic technology than many other specialities. Thus, the present review intends first to provide background on how objective kinematic parameters link with technical and non-technical skills. The potential for innovations from artificial intelligence and computer vision to track technical and nontechnical skills in real and simulated settings will then be evaluated, and last, the benefits and barriers to the uptake of such technology for the cardiothoracic community will be discussed.

Innovations in machine learning that render the analysis of real-time cardiothoracic surgery accessible even without specialist motion-tracking equipment provide a solution to understanding how the surgical environment shapes operative performance by providing objective measures of technical and co-ordinative skill in the operating theatre. Such analyses could also provide a valuable performance feedback tool for surgeons throughout their careers as they evolve or provide immediate objective indicators of factors that can impact surgical performance, such as fatigue. In terms of training, integrating such data-driven approaches with empirically validated teamwork theory may give trainees or those surgeons being mentored more substantial opportunities to develop the technical and professional skills to excel in their practice.

***Surgical expertise***

Much work has already been done to delineate markers of surgical expertise and provide measures that assess surgical skills. Whilst it is beyond the scope of the present paper to provide a comprehensive overview (reviews: [1,8]), surgical expertise is thought to comprise both technical and non-technical skills. Technical skill refers to direct psychomotor ability as governed by visuomotor aptitude, economy of movement and co-ordination [1], whereas non-technical skills encompass a broad range of abilities that support the surgical task [1,9–11]. Specifically, individual markers of expertise beyond technical skills include declarative knowledge, interpersonal skills, situational awareness, and cognitive flexibility[1].

Surgical performance depends not just on a surgeon's technical and non-technical skills but also on the skills of the surgical team and the administrative, managerial and organisational policies and procedures that support them. Thus, it is crucial to consider that surgical performance occurs within a broader context. For example, better team communication is associated with higher non-technical skill performance overall [12], and fewer miscommunications occur in teams that have a high degree of familiarity [13]. These effects on non-technical skills translate further: situational awareness in the surgical team demonstrates a strong negative relationship to the frequency of technical errors [14,15], and high team familiarity appears to be associated with lower rates of postoperative morbidity [16]. Thus, the path to cardiothoracic surgical excellence requires a holistic and integrative approach; considering context can provide insights beyond direct technical performance [9].

---

[1] Some theoretical conceptualisations of surgical skill consider cognitive skills to be distinct from non-technical skills [1].

From general cognitive research and theory into joint [17] and individual co-ordination [18], as well as applied surgical investigations [8], we know that factors beyond technical skill contribute to psychomotor performance and thus may also be observed in movement execution. Non-technical skills, group-related factors and organisational factors can all influence psychomotor performance. Consider how a cardiothoracic surgeon executes an action plan and how the surgeon must adapt their movements due to an unforeseen event such as major unexpected bleeding or a perfusion issue such as air embolism — the speed at which the surgeon adjusts quickly to new circumstances might be influenced by any number of non-technical skills (e.g. mental readiness, fatigue, anticipatory ability, cognitive flexibility or situational awareness). Such individual factors would also depend on team-level factors; for example, working with new team members may mean attention is diverted from the task towards managing new team dynamics and developing confidence and trust in those around them. If attention is directed towards managing new team dynamics, the cognitive resources available for action planning are reduced, and movement quality may suffer. Conversely, a familiar team may allow for more cognitive flexibility or ease with which to adapt to changing patient circumstances. Specifically, the surgeon does not need to monitor team members to the same extent when they understand their behavioural preferences and abilities and have an innate feeling of trust and confidence. Additionally, the tendency to monitor team members may be lower when the surgeon expects that team members will anticipate their needs [19]. Similarly, managerial or organisational level factors such as policy, established procedures or culture can influence movement by providing formalised mutual understanding among team members, which, in turn, provides a scaffolding for behaviour and minimises the cognitive resources required to coordinate.

### *Kinematic analysis of surgical performance and expertise*

A range of kinematic metrics have been shown to discriminate surgical expertise (see Table 1). For example, increased experience is associated with lower trajectory displacement during suturing, lower acceleration with non-dominant hands, and higher velocity while tying sutures [20]. Similar results have been obtained in live settings [21]. Early increases in expertise as a trainee surgeon are related to increases in psychomotor performance (e.g. increased velocity and precision of movements). In contrast, later expertise developments are characterised by physical efficiency gains [20,22,23]. Although most work in this area is not considered specialty-specific, it is important to consider the generalisability of the findings. Speaking directly to the cardiothoracic speciality, recent work indicates that the expertise effects concerning speed and physical efficiency gains are also evident whilst performing a simulated graft anastomosis [24].

| Measure | Definition for measurement | Meaning |
| --- | --- | --- |
| Idle time [20] | Length of time in which there is little movement in both hands. | May represent periods devoted to action planning or decision-making. Shorter periods of idle time are linked with higher expertise. |
| Working volume [21] | Area of space occupied by hands whilst completing a task. | Measure of efficiency with lower working volume linked with higher expertise. |
| Path length [22,23,25] | The length of the trajectory that the hand travels in a given task. | Measure of efficiency. Relative to tissue and task expertise. May be linked with either shorter or longer path lengths [25]. |
| Task completion time [24–26] | Time taken to complete a task. | Measure of efficiency with faster completion times linked with higher expertise. |

| Number of movements [24,26,27] | Number of movements used to complete a given task, sometimes within a given time. | Measure of efficiency with fewer movements linked with higher expertise. |
|---|---|---|
| Rate of change in tool orientation [25] | The speed at which the tool angle changes during movement. | Faster rates of change linked with expertise. |
| Velocity [23] | Displacement over time | Higher velocity linked with expertise. |
| Acceleration [23] | Velocity over time | Lower acceleration linked with expertise; higher acceleration may reflect attempts to make up for slower action overall. |
| Smoothness [28] | Acceleration over time | Increased smoothness linked with the expertise. Reflective of better action planning and execution. |

Table 1. Example measures of surgical expertise/performance. Note that definitions may vary slightly across studies and measurements must be considered with relevance to the task.

Cardiothoracic surgeons and surgical teams must be both experts on a technical level and expert co-ordinators. Yet, little research investigates group-level coordination during surgery using objective kinematic measurements. Recent research on general surgical trainees indicates that expertise is linked with more robust mental representations [29]. A consequence of stronger mental representations is easier and faster retrieval from memory, which can lead to better action planning and execution, as evident in the objective measurement of movement trajectory and speed of movement initiation and execution.

Research from cognitive psychology in 'joint action' also strongly emphasises the importance of mental representations. Humans tend to form joint 'internal predictive models' - models for action that team members jointly represent. These shared representations promote smooth and effective co-ordination [30,31] because they allow team members to predict the actions of co-ordinative partners, anticipate their needs and adapt their movements to accommodate those co-ordinative needs, ultimately maximising physical efficiency [32,33], and potentially cognitive efficiency [34].

Co-ordinative agents can also facilitate shared representation via communicative movements; for example, movements can be exaggerated spatially or temporally to indicate intention [35] or call attention to important parts of a movement for teaching purposes [36,37]. In attending to the movements' communicative aspects, the observer can enhance their own internal models and respond more effectively to the signaller. Kinematic analysis of surgery on the co-ordinative level will provide insight into how surgeons optimally engage co-ordinative mechanisms, including communicative movement, to facilitate and enhance task performance.

### *Marker-less movement tracking for the evaluation of surgical performance*

Kinematic analyses have historically been performed in simulated settings [38] due to the need for specialist motion-tracking equipment that would not typically be present during live cardiothoracic surgical procedures. It is possible to use traditional markered motion capture in live surgery using tools equipped with motion capture markers or by attaching them to a surgeon's hands. However, it should be noted that any changes to an instrument or a surgeon's hands may influence their movement patterns. Therefore, extensive pre-testing or experience with the new equipment should be undertaken to ensure that the introduction of markers would not impact technical performance. Further, the practice is not well adopted beyond simulated settings and minimally invasive surgery due to concerns over introducing motion capture sensors and markers into sterile environments in open surgery [39].

Assessing surgical skill in simulated settings provides only a limited picture of surgical performance, and objective measures in simulated settings cannot be directly linked to patient outcomes. Marker-less tracking techniques using custom software algorithms have demonstrated success for kinematic analysis of surgical skills in both live open surgery [40,41] and simulated settings [23,42]. Such video review is possible using only a video camera.

In computer vision, markerless motion tracking is called 'pose estimation'; here we focus on innovations in deep learning techniques. Deep learning for pose estimation involves training an artificial neural network on annotated pose data sets [43,44], so it can 'learn' to recognise poses. When provided with new videos of operations, the network identifies the poses within that video. Kinematic parameters can then be extracted using spatial and temporal data. Where deep learning requires more powerful hardware than a standard computer, the acquisition and set up of appropriate equipment for markerless motion capture is more accessible than traditional means of markered motion capture because traditional video cameras alone may be used.

Pose estimation algorithms [45–47] and toolboxes [48] have advanced considerably recently. Further, the speed with which a network can identify poses has markedly increased, making deep learning for pose estimation a viable tool in applied kinematic analysis to provide feedback to surgeons on their performance and link psychomotor metrics to patient outcomes. Such techniques can precisely track extremely delicate procedures requiring microprecision instruments, such as retinal microsurgery [49]. They thus may even be useful to analyse procedures such as coronary artery bypass surgery or in paediatric heart surgery. Further, innovations in terms of multi-person pose estimation [46] and multi-instrument pose estimation [50] provide opportunities to analyse group co-ordination, which is an under-studied area [38].

Typically, using multiple cameras to enhance 3D spatial precision is optimal. Nevertheless, cameras designed to measure depth without using makers or additional cameras have been developed—for example, Microsoft's Kinect [51]. Like some optical marker trackers, these cameras emit infrared light and read depth information from the reflected light. High-speed pose recognition can be achieved through traditional machine learning, allowing for real-time interaction [52]. Depth cameras have been used for the 3D pose estimation of medical instruments [53].

Pose estimation techniques do have limitations. For example, occluded points are not estimated or estimated with lower accuracy. Many standard methods using markers also suffer from this limitation;[2] multiple cameras in markered and markerless [56] approaches may help alleviate occlusion issues, and gap-filling techniques may be employed post hoc to deal with short durations of missing data. As an alternative to simple gap-filling techniques and to prioritise accuracy, artificial neural networks can be used to estimate the missing information [57]. Alternatively, a hybrid approach that incorporates wearable sensors may assist. Flexible sensors (accelerometers and gyroscopes) that can be unobtrusively worn under gloves may provide useful information when the line of sight is obstructed. These sensors have recently been shown to produce measures that differentiate experts and novices during a simulated graft anastomosis [24].

Training a network to  perform pose estimation adequately takes substantial computational time, often days, even with hardware with sizeable computational power. Whereas, some forms of markered motion capture are fast enough to provide feedback in milliseconds with appropriate computer hardware. Nevertheless, a pre-trained network can perform pose estimation extremely quickly to reach real-time processing speeds (10-30Hz). Last, purpose-built algorithms may be required depending on the goals of the analysis. There have been recent initiatives that aim to collate surgical tool or surgical procedure data sets [58,59], which can be used for developing pose estimation models.

Pose estimation data and other forms of kinematic data can also be fed into artificial neural networks designed to classify a cardiothoracic surgeon's skill level [60–62]. Most demonstrations have been performed in the context of robotic or laparoscopic surgery as more data is available. However, with appropriate videos, surgical skills in open cardiac or thoracic surgery could be assessed using similar approaches. To build such classifiers, neural networks are trained on data (e.g. kinematic data, videos) taken from surgeons at all skill levels. The neural network then extracts parameters common to each group and then classifies new videos based on how well they match the typical parameters of a given skill level. These parameters, however, may not always represent meaningful skill differences between levels but may be arbitrary parameters that co-occur with skill differences. Thus, cardiothoracic surgeons with unique approaches may be disadvantaged even though such technique differences may not be meaningful for patient outcomes. Further, surgical adaptations associated with environmental factors or personal characteristics may not be accounted for sufficiently. For example, classifier algorithms are known to be prejudiced as a function of their inputs and often disadvantage under-represented groups [63]. Thus, for such skill assessment algorithms to be developed, comprehensive data sets for a given procedure that demonstrate sufficient variation in action execution and skill level are needed from surgical teams and hospitals.

---

[2] Electromagnetic tracking [54] avoids issues of visual occlusion but in addition to the requirement of markers it requires that there be no magnetic interference present. Varieties of tracking techniques, and their suitability  for given tasks have been summarised elsewhere [55].

Where deep learning for kinematic analysis is not likely to raise concerns over bias because it directly measures an event, a classifier takes the decision out of human hands and can make decisional factors opaque. Thus, developers should make every effort to ensure that the training data is not biased. Ideally, any classification should be accompanied by a meaningful description of the decisional parameters (Explainable Artificial Intelligence [64]). For example, classifiers can also be designed to provide feedback on what data components are most predictive of the skill classification [60].

A recent systematic review indicated that reported machine learning methods developed to classify expertise typically achieved over 80% accuracy [65]. With further accuracy gains likely, the technology could facilitate assessment in competency-based education. With limitations in mind, we believe such methods could be most usefully employed as a formative tool that aids surgeons in developing their technical expertise, supplementing more human resource-intensive means of feedback. Indeed, a recent systematic review highlighted that personalised feedback supported by artificial Intelligence is well accepted and considered beneficial by users [66]. However, it is noted that there is a need for rigorous experimental studies that contrast traditional pedagogical interventions with those from artificial intelligence to evaluate any potential learning gains.

### *Understanding the impact of technical skill on patient outcomes via kinematic data*

Patient outcomes can be linked to technical skill. For example, higher-rated technical performance is linked with better post-operative outcomes in neonatal cardiac surgery [67]. High skill levels are also linked to operating time [20], which may consequently influence patient outcomes. For example, prolonged femoral-popliteal bypass procedures in vascular surgery are associated with increased surgical site infection and extended post-operative stays [68]. As we operate in an age of big data, opportunities to understand the factors contributing to patient outcomes within and between hospitals will become more accessible through data mining techniques. Using deep learning to tap into kinematic data is a particularly exciting innovation that will contribute even further to the standard information that is commonly extracted.

First, it is essential to look for objective measurements of expertise linked to patient outcomes in real surgery to identify the most critical aspects of technical skill. However, a necessary corollary to this work is understanding the nuance in technical expertise between surgeons. Nuanced kinematic variations [69], for example, may represent differences in technique that surgeons have adapted to their own biomechanical and cognitive needs or the surgical procedure. Further, surgeons are heavily influenced by their training, resulting in different techniques for the same procedure. Yet, international and national registry and audit data show similar outcomes between surgeons, hospitals and nations, indicating that variations in approach are not necessarily meaningful. Overall, a better understanding of which markers of technical skill are related to patient outcomes will inform cardiothoracic surgical training and development or decline over a surgeon's whole career. As with elite athletes, surgical trainees should be supported with movement feedback to explore what works for them [70]. Identifying if, how and when technical skill impacts patient outcomes will require exceptional interdisciplinary cooperation from surgeons, surgical team members, hospitals, and researchers. Large amounts of data will be required given that many other factors can influence patient outcomes within and outside the operating room.

***Understanding the routes by which nontechnical skills influence patient outcomes via kinematic data***

One route by which non-technical skills influence patient outcomes particularly pertinent to kinematic analysis is their capacity to feed into the technical execution of surgical actions [8]. For example, an overall assessment of non-technical skills [71] measuring communication and interaction, situation awareness and vigilance, co-operation and team skills, leadership and managerial skills, and decision-making was linked to the technical performance of surgeons performing carotid endarterectomy. The same is likely to be true in cardiothoracic surgery more generally.

Non-technical skills are not only a modulator of technical skills but may also influence patient outcomes directly. In fact, a video evaluation study assessing the surgical skills of surgeons (including 83 cardiac and cardiothoracic surgeons) demonstrated that increased scores for non-technical skills, independent of technical skill, were related to higher patient safety ratings [72]. This direct influence is particularly evident in crisis settings in the operating room, where non-technical skills drop considerably for all expertise levels [73] where changes in technical performance are less pronounced or negligible for highly experienced surgeons [74]. A similar pattern can be observed in response to fatigue [28]. If situational awareness drops, for example, the cardiothoracic surgeon will be less able to monitor all aspects of the surgery well and may not make the most informed patient care decisions.

Measuring non-technical skills objectively via video-based data is less straightforward than speaking to movement execution. Nevertheless, it is achievable. An investigation of nursing students showed that video-based feedback on gaze allowed the trainee to develop situational awareness [75]. In addition, recent advances show that human attention can be tracked within a task space by modelling head pose and orientation. Of course, this approach is less precise than using eye-tracking technology. However, such modelling helps understand various factors contributing to situational awareness, such as concentration loss, collaborative attention and stress levels more generally whilst engaging in collaborative tasks [76]. Further, as mentioned earlier, communicative gestures can be differentiated from goal-directed gestures [35–37] in terms of their kinematic features, and collaborative responses could be indexed by reaction times to requests. Although it is theoretically achievable to objectively measure some non-technical skills from video data as indicated by work in other fields, such an approach would need to be empirically validated within a cardiothoracic surgical setting.

Analysing co-ordinative kinematics via deep learning techniques in real crises may explain how and why performance changes. Such analyses will also provide insight into how the organisation and the wider surgical team can optimally support the operating surgeon.

### *Implications and Implementation*

Understanding markers of expertise (and learning trajectories) for various surgical tasks, how those markers relate to cognitive mechanisms that optimise performance, and subsequently, how performance impacts patient outcomes would help design targeted training programmes for cardiothoracic trainees and established  surgeons at all levels wishing to enhance their own motor expertise. Understanding the aspects of expertise linked to surgical outcomes would allow a surgeon to target the most beneficial areas of improvement at any given time during their learning curve as this changes throughout their career.

Traditional methods of teaching and development focus on repeated practice under the supervision of a more senior surgeon. With surgical coaching and mentorship, the surgeon engages in an ongoing process of performance reflection and adjustment under the guidance of a surgical coach. Although research investigating the efficacy of surgical coaching is still new [77], current evidence suggests it is highly effective for skill acquisition and development, and participants receive it well [78–80].

Deliberate individualised practice is essential to surgical skill acquisition [81] and is also an essential part of the surgical coaching process. Tracking metrics over time allows a surgeon to engage in deliberate practice by measuring improvement and using targeted feedback. Thus, kinematic tracking and feedback tools could provide further targeted guidance to complement the feedback and reflection provided in surgical coaching. For example, a user could submit a video of their own performance to software that they wish to get feedback on. The software would then analyse kinematic parameters known to reflect expertise against collective benchmarks or their own previous performance and provide targeted recommendations to improve performance. Feedback could also be provided online during simulations if such feedback was helpful.

Whilst it is important to note that engaging in facilitated reflection and problem solving with a surgical coaching tool based on artificial intelligence is unlikely to provide the same experience as with a human coach or mentor given the lack of authentic empathy and emotional intelligence, very recent research suggests good outcomes from AI coaches that aim to assist with goal attainment more generally [82]. Given that surgical training and coaching are extremely human resource intensive, increasing the availability of tools that provide opportunities to gain additional feedback without needing a human expert may assist in accelerating the development of surgical skills [83] to complement or facilitate the more resource-intensive visual assessment used in surgical coaching [84]. Furthermore, surgical coaches could use the technology to assist decision-making or feedback (human in the loop, [85]). Indeed, using audio-visual technology to review performance in coaching contexts has offered additive benefits over in-person observation alone [86].

This technology also offers the potential to monitor the surgical team to provide real-time feedback that may benefit decision-making processes during surgery. For example, given that fatigue is related to technical errors [8], when a surgical team member shows signs of fatigue during long surgeries, an alert could notify the team to take a break or alter roles. Monitoring surgical skills using pose estimation methods could also assist offline in determining how a cardiothoracic surgeon's role may need to alter toward the end of their career if objective measures of intra-operative performance begin to decline and affect patient outcomes.

Competency-based education and certification are highly labour-intensive and thus could also benefit from these technologies. Pose estimation algorithms could be used to assess if a trainee surgeon meets the competency requirements. Of course, it would be essential to assess trust and acceptance of the technology for this purpose in addition to the accuracy of the algorithm. A hybrid approach could reduce labour if trust and acceptance require human oversight (human in the loop, [85]). For example, algorithms could be used throughout training to track progression against milestones and flag when the trainee meets the competency threshold to be formally assessed by a human.

Beyond training and monitoring interventions, correlative ecological studies investigating already recorded operations could help further understand what factors link with patient outcomes. Indeed, early investigations examining intra-operative performance in laparoscopic and robotic-assisted procedures demonstrate a link to short-term patient outcomes [4]. This new understanding would provide valuable information that can be used to develop data-driven policies, procedures and environments that support the optimal performance of surgeons. By training artificial neural networks to predict and track a surgeon's movement, large-scale investigations evaluating kinematic data from recorded surgeries are possible. This research could be additionally informative for competency-based educational frameworks as there is currently no evidence to suggest that a trainee's progression through milestones links with patient outcomes [87]. By understanding more clearly what aspects of skill relate to patient outcomes, it may be possible to determine competency-based thresholds informed by empirical evidence.

Of course, implementing any performance monitoring or feedback tool should be done in such a way as to foster trust between users (cardiothoracic surgeons and their teams) and the institution. Installing cameras to monitor operations is becoming more commonplace with the use of 'operating room black boxes', but it is possible that such initiatives could result in resistance. In a cross-sectional survey of Danish healthcare professionals [88], on average, opinions toward using a black box were neutral or positive, with little concern over data safety. Conversely, in a similar study conducted in Canada [89], there were more significant concerns over data safety and the potential for litigation, highlighting the importance of considering any concerns within a societal, cultural and legislative context. Irrespective of perception, video data most often supports healthcare professionals from a legal perspective [90] and thus is more likely to offer protection than be a threat. Ultimately, the success of the technology elaborated upon within this review will rely on fostering a culture of trust and engagement with users and institutions to ensure that any concerns are addressed and there are strong institutional policies to protect and support the interests of the observed cardiothoracic surgical team.

### *Ethical and Legal Considerations*

A solid institutional policy should be developed to ensure that video footage (both from simulated and live procedures) is recorded, stored and used ethically and legally. Potential concerns and methods for addressing those concerns have been summarised in detail elsewhere [91]. It is common to raise legal fears concerning the recording and storage of footage. However, as mentioned already, these fears are likely unfounded: typically, video data protects healthcare professionals rather than puts them at risk from a legal perspective [90]. Further, it is not considered necessary for video data to be added to a patient's medical record if the video is collected solely for quality improvement because it is not in any way used for the patients care [90]. Nevertheless, there may be variations in patient consent requirements across institutions and regions. Confidentiality and anonymity should be carefully considered as, in some cases, the nature of the research would require analysis of identifiable or sensitive personal information.

### *Conclusion*

New means of analysing surgical performance open doors to understanding surgical excellence in the cardiothoracic specialty. Other disciplines have traditionally benefitted from technological innovations around training and the objective measurement of performance; the fields of computer vision and Artificial Intelligence now offer opportunities that are ideal for use in the cardiothoracic surgical environment.  Further, these tools are feasible to use within in the operating room which will assist in understanding how technical and non-technical skills influence patient outcomes. However, the technology is still in the early stages, and thus, further innovation will require commitment and partnership from hospitals and cardiothoracic surgeons to provide (1) data that can be used to develop feedback tools and (2) constructive direction to ensure that any tool used for applied purposes has been developed to meet the needs of the user adequately. With data from simulated and real surgical settings, research aimed at understanding how expertise relates to the cognitive mechanisms that support psychomotor performance within the context of surgery will further help design targeted training interventions and surgical environments more optimally to enhance surgical outcomes. With big data generated across many institutions, it may be possible to develop data-driven guidelines for task execution, and team coordination that reduce the surgeon and team's physical and cognitive load [6].

**CRediT Statement**

Conceptualisation: SC, MDC. Writing – Original Draft: MDC. Writing – Revising and Editing: SC, HPHS, MDC.

**Declarations**

*Ethics approval and consent to participate:* Not applicable

*Consent for publication:* Not applicable

*Availability of data and materials:* Not applicable

*Competing Interests:* The authors declare that they have no competing interests

*Funding:* None

*Authors contributions:* Conceptualisation: SC, MDC. Writing – Original Draft: MDC. Writing – Revising and Editing: SC, HPHS, MDC. All authors read and approved the final manuscript.

*Acknowledgements:* None

References

[1] Madani A, Vassiliou MC, Watanabe Y, Al-Halabi B, Al-Rowais MS, Deckelbaum DL, et al. What Are the Principles That Guide Behaviors in the Operating Room?: Creating a Framework to Define and Measure Performance. Annals of Surgery 2017;265:255–67. https://doi.org/10.1097/SLA.0000000000001962.

[2] Petancevski EL, Inns J, Fransen J, Impellizzeri FM. The effect of augmented feedback on the performance and learning of gross motor and sport-specific skills: A systematic review. Psychology of Sport and Exercise 2022;63:102277. https://doi.org/10.1016/j.psychsport.2022.102277.

[3] Yule S, Janda A, Likosky DS. Surgical Sabermetrics: Applying Athletics Data Science to Enhance Operative Performance. Annals of Surgery Open 2021;2:e054. https://doi.org/10.1097/AS9.0000000000000054.

[4] Balvardi S, Kammili A, Hanson M, Mueller C, Vassiliou M, Lee L, et al. The association between video-based assessment of intraoperative technical performance and patient outcomes: a systematic review. Surg Endosc 2022;36:7938–48. https://doi.org/10.1007/s00464-022-09296-6.

[5] Mazer L, Varban O, Montgomery JR, Awad MM, Schulman A. Video is better: why aren't we using it? A mixed-methods study of the barriers to routine procedural video recording and case review. Surg Endosc 2022;36:1090–7. https://doi.org/10.1007/s00464-021-08375-4.

[6] Chadebecq F, Vasconcelos F, Mazomenos E, Stoyanov D. Computer Vision in the Surgical Operating Room. VIS 2020;36:456–62. https://doi.org/10.1159/000511934.

[7] Vedula SS, Hager GD. Surgical data science: the new knowledge domain. Innov Surg Sci 2017;2:109–21. https://doi.org/10.1515/iss-2017-0004.

[8] Hull L, Arora S, Aggarwal R, Darzi A, Vincent C, Sevdalis N. The Impact of Nontechnical Skills on Technical Performance in Surgery: A Systematic Review. Journal of the American College of Surgeons 2012;214:214–30. https://doi.org/10.1016/j.jamcollsurg.2011.10.016.

[9] Carthey J, de Leval MR, Wright DJ, Farewell VT, Reason JT. Behavioural markers of surgical excellence. Safety Science 2003;41:409–25. https://doi.org/10.1016/S0925-7535(01)00076-5.

[10] Crossley J, Marriott J, Purdie H, Beard JD. Prospective observational study to evaluate NOTSS (Non-Technical Skills for Surgeons) for assessing trainees' non-technical performance in the operating theatre. Br J Surg 2011;98:1010–20. https://doi.org/10.1002/bjs.7478.

[11] Yule S, Flin R, Paterson-Brown S, Maran N. Non-technical skills for surgeons in the operating room: A review of the literature. Surgery 2006;139:140–9. https://doi.org/10.1016/j.surg.2005.06.017.

[12] Gillespie BM, Harbeck E, Kang E, Steel C, Fairweather N, Chaboyer W. Correlates of non-technical skills in surgery: a prospective study. BMJ Open 2017;7:e014480. https://doi.org/10.1136/bmjopen-2016-014480.

[13] Gillespie BM, Chaboyer W, Fairweather N. Interruptions and miscommunications in surgery: an observational study. AORN J 2012;95:576–90. https://doi.org/10.1016/j.aorn.2012.02.012.

[14] Siu J, Maran N, Paterson-Brown S. Observation of behavioural markers of non-technical skills in the operating room and their relationship to intra-operative incidents. The Surgeon 2016;14:119–28. https://doi.org/10.1016/j.surge.2014.06.005.

[15] McCulloch P, Mishra A, Handa A, Dale T, Hirst G, Catchpole K. The effects of aviation-style non-technical skills training on technical performance and outcome in the operating theatre. Qual Saf Health Care 2009;18:109–15. https://doi.org/10.1136/qshc.2008.032045.

[16] Kurmann A, Keller S, Tschan-Semmer F, Seelandt J, Semmer NK, Candinas D, et al. Impact of team familiarity in the operating room on surgical complications. World J Surg 2014;38:3047–52. https://doi.org/10.1007/s00268-014-2680-2.

[17] Sebanz N, Knoblich G. Progress in Joint-Action Research. Curr Dir Psychol Sci 2021:0963721420984425. https://doi.org/10.1177/0963721420984425.

[18] Jeannerod M. Motor Cognition: What Actions Tell the Self. OUP Oxford; 2006.

[19] Frasier LL, Pavuluri Quamme SR, Ma Y, Wiegmann D, Leverson G, DuGoff EH, et al. Familiarity and Communication in the Operating Room. Journal of Surgical Research 2019;235:395–403. https://doi.org/10.1016/j.jss.2018.09.079.

[20] D'Angelo A-LD, Rutherford DN, Ray RD, Laufer S, Kwan C, Cohen ER, et al. Idle time: an underdeveloped performance metric for assessing surgical skill. The American Journal of Surgery 2015;209:645–51. https://doi.org/10.1016/j.amjsurg.2014.12.013.

[21] D'Angelo A-LD, Rutherford DN, Ray RD, Laufer S, Mason A, Pugh CM. Working volume: validity evidence for a motion-based metric of surgical efficiency. The American Journal of Surgery 2016;211:445–50. https://doi.org/10.1016/j.amjsurg.2015.10.005.

[22] Glarner CE, Hu Y-Y, Chen C-H, Radwin RG, Zhao Q, Craven MW, et al. Quantifying technical skills during open operations using video-based motion analysis. Surgery 2014;156:729–34. https://doi.org/10.1016/j.surg.2014.04.054.

[23] Azari D, Miller BL, Le BV, Greenberg CC, Radwin RG. Quantifying surgeon maneuevers across experience levels through marker-less hand motion kinematics of simulated surgical tasks. Applied Ergonomics 2020;87:103136. https://doi.org/10.1016/j.apergo.2020.103136.

[24] Boyajian GP, Zulbaran-Rojas A, Najafi B, Atique MdMU, Loor G, Gilani R, et al. Development of a Sensor Technology to Objectively Measure Dexterity for Cardiac Surgical Proficiency. The Annals of Thoracic Surgery 2023. https://doi.org/10.1016/j.athoracsur.2023.07.013.

[25] Sharon Y, Jarc AM, Lendvay TS, Nisky I. Rate of Orientation Change as a New Metric for Robot-Assisted and Open Surgical Skill Evaluation. IEEE Transactions on Medical Robotics and Bionics 2021;3:414–25. https://doi.org/10.1109/TMRB.2021.3073209.

[26] Bann SD, Khan MS, Darzi AW. Measurement of Surgical Dexterity Using Motion Analysis of Simple Bench Tasks. World J Surg 2003;27:390–4. https://doi.org/10.1007/s00268-002-6769-7.

[27] Xeroulis GJ, Park J, Moulton C-A, Reznick RK, Leblanc V, Dubrowski A. Teaching suturing and knot-tying skills to medical students: a randomized controlled study comparing computer-based video instruction and (concurrent and summary) expert feedback. Surgery 2007;141:442–9. https://doi.org/10.1016/j.surg.2006.09.012.

[28] Kahol K, Leyba MJ, Deka M, Deka V, Mayes S, Smith M, et al. Effect of fatigue on psychomotor and cognitive skills. The American Journal of Surgery 2008;195:195–204. https://doi.org/10.1016/j.amjsurg.2007.10.004.

[29] Yeh VJ-H, Mukhtar F, Yudkowsky R, Baloul MS, Farley DR, Cook DA. Response Process Validity Evidence for Video Commentary Assessment in Surgery: A Qualitative Study. Journal of Surgical Education 2022;79:1270–81. https://doi.org/10.1016/j.jsurg.2022.05.006.

[30] Curioni A, Vesper C, Knoblich G, Sebanz N. Reciprocal information flow and role distribution support joint action coordination. Cognition 2019;187:21–31. https://doi.org/10.1016/j.cognition.2019.02.006.

[31] Wolpert DM, Doya K, Kawato M. A unifying computational framework for motor control and social interaction. Philos Trans R Soc Lond B Biol Sci 2003;358:593–602. https://doi.org/10.1098/rstb.2002.1238.

[32] Török G, Pomiechowska B, Csibra G, Sebanz N. Rationality in Joint Action: Maximizing Coefficiency in Coordination. Psychol Sci 2019;30:930–41. https://doi.org/10.1177/0956797619842550.

[33] Constable MD, Bayliss AP, Tipper SP, Spaniol AP, Pratt J, Welsh TN. Ownership status influences the degree of joint facilitatory behavior. Psychological Science 2016;27:1371–8. https://doi.org/10.1177/0956797616661544.

[34] Lagomarsino M, Lorenzini M, Constable MD, De Momi E, Becchio C, Ajoudani A. Maximising Coefficiency of Human-Robot Handovers Through Reinforcement Learning. IEEE Robotics and Automation Letters 2023;8:4378–85. https://doi.org/10.1109/LRA.2023.3280752.

[35] Pezzulo G, Donnarumma F, Dindo H. Human sensorimotor communication: A theory of signaling in online social interactions. PLoS ONE 2013;8:e79876. https://doi.org/10.1371/journal.pone.0079876.

[36] Strachan JWA, Curioni A, Constable MD, Charbonneu M. A methodology for distinguishing copying and reconstruction in cultural transmission episodes. 42nd Annual Meeting of the Cognitive, 2020.

[37] Strachan JWA, Curioni A, Constable MD, Knoblich G, Charbonneau M. Evaluating the relative contributions of copying and reconstruction processes in cultural transmission episodes. PLOS ONE 2021;16:e0256901. https://doi.org/10.1371/journal.pone.0256901.

[38] Mitchell EL, Arora S, Moneta GL, Kret MR, Dargon PT, Landry GJ, et al. A systematic review of assessment of skill acquisition and operative competency in vascular surgical training. Journal of Vascular Surgery 2014;59:1440–55. https://doi.org/10.1016/j.jvs.2014.02.018.

[39] Reiley CE, Lin HC, Yuh DD, Hager GD. Review of methods for objective surgical skill evaluation. Surg Endosc 2011;25:356–66. https://doi.org/10.1007/s00464-010-1190-z.

[40] Frasier LL, Azari DP, Ma Y, Quamme SRP, Radwin RG, Pugh CM, et al. A marker-less technique for measuring kinematics in the operating room. Surgery 2016;160:1400–13. https://doi.org/10.1016/j.surg.2016.05.004.

[41] Kadkhodamohammadi A, Gangi A, de Mathelin M, Padoy N. A Multi-view RGB-D Approach for Human Pose Estimation in Operating Rooms. 2017 IEEE Winter Conference on Applications of Computer Vision (WACV), 2017, p. 363–72. https://doi.org/10.1109/WACV.2017.47.

[42] Casy T, Tronchot A, Thomazeau H, Morandi X, Jannin P, Huaulmé A. "Stand-up straight!": human pose estimation to evaluate postural skills during orthopedic surgery simulations. Int J Comput Assist Radiol Surg 2023;18:279–88. https://doi.org/10.1007/s11548-022-02762-5.

[43] Ionescu C, Papava D, Olaru V, Sminchisescu C. Human3.6M: Large Scale Datasets and Predictive Methods for 3D Human Sensing in Natural Environments. IEEE Trans Pattern Anal Mach Intell 2014;36:1325–39. https://doi.org/10.1109/TPAMI.2013.248.

[44] Lin T-Y, Maire M, Belongie S, Hays J, Perona P, Ramanan D, et al. Microsoft COCO: Common Objects in Context. In: Fleet D, Pajdla T, Schiele B, Tuytelaars T, editors. Computer Vision – ECCV 2014, vol. 8693, Cham: Springer International Publishing; 2014, p. 740–55. https://doi.org/10.1007/978-3-319-10602-1_48.

[45] Cao Z, Simon T, Wei S-E, Sheikh Y. Realtime Multi-person 2D Pose Estimation Using Part Affinity Fields. 2017 IEEE Conference on Computer Vision and Pattern Recognition (CVPR), Honolulu, HI: IEEE; 2017, p. 1302–10. https://doi.org/10.1109/CVPR.2017.143.

[46] Huang Y, Shum HPH, Ho ESL, Aslam N. High-speed multi-person pose estimation with deep feature transfer. Computer Vision and Image Understanding 2020;197–198:103010. https://doi.org/10.1016/j.cviu.2020.103010.

[47] Insafutdinov E, Pishchulin L, Andres B, Andriluka M, Schiele B. DeeperCut: A Deeper, Stronger, and Faster Multi-person Pose Estimation Model. In: Leibe B, Matas J, Sebe N, Welling M, editors. Computer Vision – ECCV 2016, Cham: Springer International Publishing; 2016, p. 34–50. https://doi.org/10.1007/978-3-319-46466-4_3.

[48] Mathis A, Mamidanna P, Cury KM, Abe T, Murthy VN, Mathis MW, et al. DeepLabCut: markerless pose estimation of user-defined body parts with deep learning. Nature Neuroscience 2018;21:1281–9. https://doi.org/10.1038/s41593-018-0209-y.

[49] Alsheakhali M, Eslami A, Roodaki H, Navab N. CRF-Based Model for Instrument Detection and Pose Estimation in Retinal Microsurgery. Computational and Mathematical Methods in Medicine 2016;2016:1–10. https://doi.org/10.1155/2016/1067509.

[50] Kurmann T, Marquez Neila P, Du X, Fua P, Stoyanov D, Wolf S, et al. Simultaneous Recognition and Pose Estimation of Instruments in Minimally Invasive Surgery. In: Descoteaux M, Maier-Hein L, Franz A, Jannin P, Collins DL, Duchesne S, editors. Medical Image Computing and Computer-Assisted Intervention – MICCAI 2017, Cham: Springer International Publishing; 2017, p. 505–13. https://doi.org/10.1007/978-3-319-66185-8_57.

[51]  Microsoft Corp. Kinect for Xbox 360 n.d.

[52]  Shotton J, Fitzgibbon A, Cook M, Sharp T, Finocchio M, Moore R, et al. Real-time human pose recognition in parts from single depth images. CVPR 2011, 2011, p. 1297–304. https://doi.org/10.1109/CVPR.2011.5995316.

[53]  Liu J, Tateyama T, Iwamoto Y, Chen Y-W. A Preliminary Study of Kinect-Based Real-Time Hand Gesture Interaction Systems for Touchless Visualizations of Hepatic Structures in Surgery. 医用画像情報学会雑誌 2019;36:128–35. https://doi.org/10.11318/mii.36.128.

[54]  Polhemus. Polhemus n.d.

[55]  Rutherford DN, D'Angelo A-LD, Law KE, Pugh CM. Advanced Engineering Technology for Measuring Performance. Surgical Clinics of North America 2015;95:813–26. https://doi.org/10.1016/j.suc.2015.04.005.

[56]  Kocabas M, Karagoz S, Akbas E. Self-Supervised Learning of 3D Human Pose Using Multi-View Geometry. 2019 IEEE/CVF Conference on Computer Vision and Pattern Recognition (CVPR), Long Beach, CA, USA: IEEE; 2019, p. 1077–86. https://doi.org/10.1109/CVPR.2019.00117.

[57]  Kanazawa A, Black MJ, Jacobs DW, Malik J. End-to-End Recovery of Human Shape and Pose. 2018 IEEE/CVF Conference on Computer Vision and Pattern Recognition, 2018, p. 7122–31. https://doi.org/10.1109/CVPR.2018.00744.

[58]  Bouget D, Allan M, Stoyanov D, Jannin P. Vision-based and marker-less surgical tool detection and tracking: a review of the literature. Medical Image Analysis 2017;35:633–54. https://doi.org/10.1016/j.media.2016.09.003.

[59]  Srivastav V, Issenhuth T, Kadkhodamohammadi A, de Mathelin M, Gangi A, Padoy N. MVOR: A Multi-view RGB-D Operating Room Dataset for 2D and 3D Human Pose Estimation. arXivOrg 2018. https://arxiv.org/abs/1808.08180v3 (accessed October 8, 2023).

[60]  Ismail Fawaz H, Forestier G, Weber J, Idoumghar L, Muller P-A. Evaluating Surgical Skills from Kinematic Data Using Convolutional Neural Networks. In: Frangi AF, Schnabel JA, Davatzikos C, Alberola-López C, Fichtinger G, editors. Medical Image Computing and Computer Assisted Intervention – MICCAI 2018, Cham: Springer International Publishing; 2018, p. 214–21. https://doi.org/10.1007/978-3-030-00937-3_25.

[61]  Khalid S, Goldenberg M, Grantcharov T, Taati B, Rudzicz F. Evaluation of Deep Learning Models for Identifying Surgical Actions and Measuring Performance. JAMA Network Open 2020;3:e201664. https://doi.org/10.1001/jamanetworkopen.2020.1664.

[62]  Levin M, McKechnie T, Khalid S, Grantcharov TP, Goldenberg M. Automated Methods of Technical Skill Assessment in Surgery: A Systematic Review. Journal of Surgical Education 2019;76:1629–39. https://doi.org/10.1016/j.jsurg.2019.06.011.

[63]  Holstein K, Wortman Vaughan J, Daumé H, Dudik M, Wallach H. Improving Fairness in Machine Learning Systems: What Do Industry Practitioners Need? Proceedings of the 2019 CHI Conference on Human Factors in Computing Systems, New York, NY, USA: Association for Computing Machinery; 2019, p. 1–16. https://doi.org/10.1145/3290605.3300830.

[64]  Taylor JET, Taylor GW. Artificial cognition: How experimental psychology can help generate explainable artificial intelligence. Psychon Bull Rev 2020. https://doi.org/10.3758/s13423-020-01825-5.

[65]  Lam K, Chen J, Wang Z, Iqbal FM, Darzi A, Lo B, et al. Machine learning for technical skill assessment in surgery: a systematic review. Npj Digit Med 2022;5:1–16. https://doi.org/10.1038/s41746-022-00566-0.

[66]  Kirubarajan A, Young D, Khan S, Crasto N, Sobel M, Sussman D. Artificial Intelligence and Surgical Education: A Systematic Scoping Review of Interventions. Journal of Surgical Education 2022;79:500–15. https://doi.org/10.1016/j.jsurg.2021.09.012.

[67]  Nathan M, Karamichalis JM, Liu H, del Nido P, Pigula F, Thiagarajan R, et al. Intraoperative adverse events can be compensated by technical performance in neonates and infants after cardiac surgery: A prospective study. The Journal of Thoracic and Cardiovascular Surgery 2011;142:1098-1107.e5. https://doi.org/10.1016/j.jtcvs.2011.07.003.

[68] Tan T-W, Kalish JA, Hamburg NM, Rybin D, Doros G, Eberhardt RT, et al. Shorter duration of femoral-popliteal bypass is associated with decreased surgical site infection and shorter hospital length of stay. J Am Coll Surg 2012;215:512–8. https://doi.org/10.1016/j.jamcollsurg.2012.06.007.

[69] Datta V, Mackay S, Mandalia M, Darzi A. The use of electromagnetic motion tracking analysis to objectively measure open surgical skill in the laboratory-based model. J Am Coll Surg 2001;193:479–85. https://doi.org/10.1016/s1072-7515(01)01041-9.

[70] Glazier PS. Beyond animated skeletons: How can biomechanical feedback be used to enhance sports performance? Journal of Biomechanics 2021;129:110686. https://doi.org/10.1016/j.jbiomech.2021.110686.

[71] Sevdalis N, Davis R, Koutantji M, Undre S, Darzi A, Vincent CA. Reliability of a revised NOTECHS scale for use in surgical teams. Am J Surg 2008;196:184–90. https://doi.org/10.1016/j.amjsurg.2007.08.070.

[72] Yule S, Gupta A, Gazarian D, Geraghty A, Smink DS, Beard J, et al. Construct and criterion validity testing of the Non-Technical Skills for Surgeons (NOTSS) behaviour assessment tool using videos of simulated operations. Br J Surg 2018;105:719–27. https://doi.org/10.1002/bjs.10779.

[73] Black SA, Nestel DF, Kneebone RL, Wolfe JHN. Assessment of surgical competence at carotid endarterectomy under local anaesthesia in a simulated operating theatre. British Journal of Surgery 2010;97:511–6. https://doi.org/10.1002/bjs.6938.

[74] Wetzel CM, Black SA, Hanna GB, Athanasiou T, Kneebone RL, Nestel D, et al. The Effects of Stress and Coping on Surgical Performance During Simulations. Annals of Surgery 2010;251:171–6. https://doi.org/10.1097/SLA.0b013e3181b3b2be.

[75] O'Meara P, Munro G, Williams B, Cooper S, Bogossian F, Ross L, et al. Developing situation awareness amongst nursing and paramedicine students utilizing eye tracking technology and video debriefing techniques: A proof of concept paper. International Emergency Nursing 2015;23:94–9. https://doi.org/10.1016/j.ienj.2014.11.001.

[76] Lagomarsino M, Lorenzini M, Balatti P, Momi ED, Ajoudani A. Pick the Right Co-Worker: Online Assessment of Cognitive Ergonomics in Human-Robot Collaborative Assembly. IEEE Trans Cogn Dev Syst 2022:1–1. https://doi.org/10.1109/TCDS.2022.3182811.

[77] Skinner SC, Mazza S, Carty MJ, Lifante J-C, Duclos A. Coaching for Surgeons: A Scoping Review of the Quantitative Evidence. Ann Surg Open 2022;3:e179. https://doi.org/10.1097/AS9.0000000000000179.

[78] Bonrath EM, Dedy NJ, Gordon LE, Grantcharov TP. Comprehensive Surgical Coaching Enhances Surgical Skill in the Operating Room: A Randomized Controlled Trial. Annals of Surgery 2015;262:205–12. https://doi.org/10.1097/SLA.0000000000001214.

[79] Gagnon L-H, Abbasi N. Systematic review of randomized controlled trials on the role of coaching in surgery to improve learner outcomes. The American Journal of Surgery 2018;216:140–6. https://doi.org/10.1016/j.amjsurg.2017.05.003.

[80] Greenberg CC, Byrnes ME, Engler TA, Quamme SP, Thumma JR, Dimick JB. Association of a Statewide Surgical Coaching Program with Clinical Outcomes and Surgeon Perceptions. Ann Surg 2021;273:1034–9. https://doi.org/10.1097/SLA.0000000000004800.

[81] Issenberg SB, McGaghie WC, Petrusa ER, Lee Gordon D, Scalese RJ. Features and uses of high-fidelity medical simulations that lead to effective learning: a BEME systematic review. Med Teach 2005;27:10–28. https://doi.org/10.1080/01421590500046924.

[82] Terblanche N, Molyn J, Haan E de, Nilsson VO. Comparing artificial intelligence and human coaching goal attainment efficacy. PLOS ONE 2022;17:e0270255. https://doi.org/10.1371/journal.pone.0270255.

[83] Safir O, Williams CK, Dubrowski A, Backstein D, Carnahan H. Self-directed practice schedule enhances learning of suturing skills. Can J Surg 2013;56:E142–7. https://doi.org/10.1503/cjs.019512.

[84] Hu Y-Y, Peyre SE, Arriaga AF, Osteen RT, Corso KA, Weiser TG, et al. Post Game Analysis: Using Video-Based Coaching for Continuous Professional Development. J Am Coll Surg 2012;214:115–24. https://doi.org/10.1016/j.jamcollsurg.2011.10.009.

[85] Enarsson T, Enqvist L, Naarttijärvi M. Approaching the human in the loop – legal perspectives on hybrid human/algorithmic decision-making in three contexts. Information & Communications Technology Law 2022;31:123–53. https://doi.org/10.1080/13600834.2021.1958860.

[86] Gunn EGM, Ambler OC, Nallapati SC, Smink DS, Tambyraja AL, Yule S. Coaching with audiovisual technology in acute-care hospital settings: systematic review. BJS Open 2023;7:zrad017. https://doi.org/10.1093/bjsopen/zrad017.

[87] Kendrick DE, Thelen AE, Chen X, Gupta T, Yamazaki K, Krumm AE, et al. Association of Surgical Resident Competency Ratings With Patient Outcomes. Acad Med 2023;98:813–20. https://doi.org/10.1097/ACM.0000000000005157.

[88] Strandbygaard J, Dose N, Moeller KE, Gordon L, Shore E, Rosthøj S, et al. Healthcare professionals' perception of safety culture and the Operating Room (OR) Black Box technology before clinical implementation: a cross-sectional survey. BMJ Open Qual 2022;11:e001819. https://doi.org/10.1136/bmjoq-2022-001819.

[89] Gordon L, Reed C, Sorensen JL, Schulthess P, Strandbygaard J, Mcloone M, et al. Perceptions of safety culture and recording in the operating room: understanding barriers to video data capture. Surg Endosc 2022;36:3789–97. https://doi.org/10.1007/s00464-021-08695-5.

[90] van Dalen ASHM, Legemaate J, Schlack WS, Legemate DA, Schijven MP. Legal perspectives on black box recording devices in the operating environment. BJS (British Journal of Surgery) 2019;106:1433–41. https://doi.org/10.1002/bjs.11198.

[91] Xiao Y, Schimpff S, Mackenzie C, Merrell R, Entin E, Voigt R, et al. Video Technology to Advance Safety in the Operating Room and Perioperative Environment. Surg Innov 2007;14:52–61. https://doi.org/10.1177/1553350607299777.